\documentclass[letterpaper]{article}
\usepackage{aaai}
\usepackage{times}
\usepackage[small,compact]{titlesec}
\usepackage{helvet}
\usepackage{courier}
\usepackage{color}
\usepackage{amsmath,amssymb}
\usepackage{multirow}
\usepackage{graphicx}
\usepackage{booktabs}
\usepackage{hyperref}
\usepackage{float}
\usepackage{url}
\usepackage{breakurl}
\usepackage{enumerate}
\usepackage{named}
\usepackage{amssymb}

\usepackage{algpseudocode}
\usepackage{algorithm}
\newcommand{\comment}[1]{}
\newcommand{\mytilde}{\raise.17ex\hbox{$\scriptstyle\mathtt{\sim}$}}

\definecolor{light-gray}{gray}{0.4}
\begin{document}
%
\title{Entity-Specific Sentiment Classification of Yahoo News Comments}
\author{Prakhar Biyani$^1$, Cornelia Caragea$^2$ and Narayan Bhamidipati$^1$\\
$^1$Yahoo Labs, Sunnyvale, California, USA\\
$^2$Computer Science and Engineering, University of North Texas, Denton, Texas, USA \\
Email: pxb5080@yahoo-inc.com.com, ccaragea@unt.edu, narayanb@yahoo-inc.com
}

\maketitle

\begin{abstract}
\vspace{-2mm}
Sentiment classification is widely used for product reviews and in online social media such as forums, Twitter, and blogs. However, the problem of classifying the sentiment of user comments on news sites has not been addressed yet. News sites cover a wide range of domains including politics, sports, technology, and entertainment, in contrast to other online social sites such as forums and review sites, which are specific to a particular domain. A user associated with a news site is likely to post comments on diverse topics (e.g., politics, smartphones, and sports) or diverse entities (e.g., Obama, iPhone, or Google). Classifying the sentiment of users tied to various entities may help obtain a {\em holistic} view of their personality, which could be useful in applications such as online advertising, content personalization, and political campaign planning. In this paper, we formulate the problem of entity-specific sentiment classification of comments posted on news articles in Yahoo News and propose novel features that are specific to news comments. Experimental results show that our models outperform state-of-the-art baselines. 
\end{abstract}

\comment{
\vspace{-6mm}
\terms{Human factors, Algorithms, Experimentation.}
\vspace{-3mm}
\\
\vspace{-3mm}
}
\vspace{-3.5mm}
\section{Introduction}
\label{intro}
Online news aggregator sites such as Yahoo News are a place for users to get in touch with developments across various domains. 
In addition to reading news articles, users post comments giving their opinions/sentiments about the topics or entities discussed in the news articles, while interacting (agreeing or disagreeing) with other users. 
This has resulted in vast amounts of User Generated Content in the form of user comments. An interesting characteristic of news sites is that they cover a wide range of domains such as politics, sports, technology, and entertainment,  in contrast to other online social sites, including forums (e.g., UbuntuForums and TripAdvisor) and review sites (e.g., dpreview.com for digital cameras and notebookreview.com for laptops), which are specific to a particular domain. Hence, the activity of a user in terms of posting comments is potentially much more diverse in news sites as compared to other social platforms.

Although it is not uncommon for users to make general comments/statements on various topics or to comment on unrelated entities that they like or dislike, in many cases, comments on a news article contain the sentiments of users tied to specific entities in the article (e.g., Obama or Android). 
Classifying the sentiments of a particular user on diverse entities may help obtain a {\em holistic} view of their personality\footnote{In adherence to Yahoo's privacy policy, all user activity is anonymized and the actual user's identity is unknown to us.}. 
For example, the sentiments of a user's comments on news articles tied to specific entities  related to politics, smartphones and online retail may help infer her political orientation, preference for a particular mobile operating system (Android vs. iOS) and liking of a particular online retailer (Walmart vs. Target). 
User sentiments across articles on an entity (e.g., iPhone) can also be followed to determine how sentiments evolve or change over time, and what factors can cause the sentiment change. 
Analyzing the sentiment of these user comments 
can help understand the {\em user} better which, in turn, can be used to provide greater personalization and improve serving targeted ads to those users.

However, despite the evidence of strong value in analyzing the sentiment of users tied to specific entities, there have not been any reported works on this problem. The problem of identifying the sentiment polarity of these comments remains inherently difficult due to several main challenges, including irrelevant entities and implicit sentiment.

{\bf Irrelevant entities:} Comments often have entities that are not important with respect to sentiment analysis. Let us consider the following example:	

\vspace{1mm}
\noindent {\bf Example 1:} {\em Great!} Foxnews {\em poll:} Obama +9; CNN poll: Obama +7; Reuters/Ipsos poll: Obama +9.  {\em I feel a landslide in the making. Gobama! Gobama! Gobama!}

In this example, the commenter has a positive sentiment for Obama and no sentiment for entities Foxnews, CNN, Reuters and Ipsos, which are irrelevant for sentiment analysis. Unlike other domains such as product reviews where the sentiment is expressed towards a precisely defined target (i.e., a product or its features), known beforehand, in our domain, the set of entities is not known a priori and covers a wide range of entities, with many of them being irrelevant. In the example above, a traditional sentiment classifier would possibly identify the sentiment for Foxnews as positive due to its close proximity with the sentiment clue ``Great!'', leading to inaccurate results. 

{\bf Implicit sentiment}: Users often express sentiments implicitly in their comments by using ironies, analogies and rhetoric, making it hard to detect the sentiment towards entities \cite{gonzalez2011identifying,utsumi2000verbal}. Let us consider the following examples: 
		
\vspace{1mm}
\noindent {\bf Example 2:} {\em I've heard that} Hillary Clinton {\em modeled herself after Nurse Ratched}.
	
\vspace{1mm}	
\noindent {\bf Example 3:} {\em Who on earth would even buy} Facebook {\em stock?}

\vspace{1mm}
		The first example has a negative sentiment about Hillary Clinton expressed through the analogy with ``Nurse Ratched'', who is a negative fictional character. The second example is a rhetorical question expressing a negative sentiment about Facebook. Typical sentiment classification approaches would label these examples as neutral due to the lack of sentiment clues \cite{ding2008holistic,feelBetter,zhang2011combining,meng2012entity}.

Against this background, one question that can be raised is: {\em Can we design techniques to effectively identify and filter out irrelevant entities in news comments and further perform accurate sentiment classification of entities for which a sentiment is expressed?} The research that we describe in this paper addresses specifically this question.

{\bf Contributions.} We address the problem of entity-specific sentiment analysis. More precisely, we formulate the problem as a two-stage binary classification. First, we identify entities that are relevant with respect to sentiment analysis, while filtering out irrelevant entities. Second, we classify the sentiment expressed towards {\em relevant} entities as positive or negative. Although there are several works on analyzing sentiments of news articles, the current problem is significantly different (as detailed in Section 2). To the best of our knowledge, there are no reported works on this problem.
The contributions of our work are as follows:
	
	1. We propose an approach for {\em context extraction} of entities discussed in news comments and show that it substantially improves sentiment classification.
	
	2. We design {\em novel features} for both classification tasks above. Specifically, we design: (1) non-lexical features for identifying relevant entities and show that these features are more informative than the lexicon-based features and the ``bag-of-words''  used in previous works on subjectivity analysis; (2) comment-specific features for sentiment classification of entities in comments. 
	
	3. We show experimentally that our sentiment classifiers trained using the proposed features extracted from the entity-specific contexts outperform several state-of-the-art approaches to sentiment classification. 

	

\vspace{-1mm}
\section{Related Work}
\label{related}
\vspace{-1.5mm}
Sentiment analysis (SA) is widely researched due to its important applications in mining, analyzing and summarizing user opinions in online product reviews~\cite{sentiWords,ly2011product,ding2008holistic}. 
Here, we review some of the relevant sentiment analysis works.

\textbf{Entity-independent SA (EISA):} EISA deals with identifying sentiment of a text without linking the sentiment to an entity for which it is expressed. 
EISA is mainly researched in the domain of product reviews, where a review is assumed to contain sentiments about a particular product and, hence, the linking is not required~\cite{supervised1,supervised2,McDonald2007,Wan2009,Li2012a}. 
Pang et al.~\cite{supervised1} used supervised machine learning algorithms trained on lexical and syntactic features such as unigrams, bigrams and POS tags, for sentiment analysis of movie reviews. 
In their later work, they improve the sentiment classification by considering only the subjective sentences and applying polarity classifiers (developed in their previous work) on those sentences~\cite{supervised2}.  
Wan et al.~\cite{Wan2009} use co-training for sentiment classification of Chinese product reviews. 
They use machine translation to obtain the training data from labeled English reviews. 
For a Chinese review, its Chinese features and the translated English features represent the two independent views that are used in  co-training.

\textbf{Entity-dependent SA (EDSA)}:
EDSA, on the other hand, links sentiment to its target entity~\cite{ding2008holistic,Nasukawa,engonopoulos2011els,zhang2011combining,meng2012entity}. 
Ding et al.~\shortcite{ding2008holistic} performed EDSA on product reviews using a lexicon-based approach. For an entity, they calculated its sentiment score by adding sentiment orientation $(\pm1)$ of opinion words co-occurring with the entity in a sentence. 
Meng et al.~\shortcite{meng2012entity} used a similar approach for sentiment classification of tweets and determine sentiment orientation by aggregating sentiments of opinion words. 
In contrast, we use supervised learning models built using several newly designed features in addition to lexicon-based features. The lexicon-based approach is one of our baselines. 

\textbf{SA in News Sites}: There are several works on sentiment classification of news articles~\cite{godbole2007large,devitt2007sentiment}. However, sentiment classification of news comments is a much more difficult task compared to that of news articles since, unlike news articles, news comments are short, noisy, incoherent, and comprise of very informal writing styles. We found a few works focusing on news comments for analyzing their quality of discourse~\cite{newsComments1} and diversifying them for presenting a comprehensive view of news articles to the readers~\cite{newsComments2}. However, these works are different from ours in nature. 

\vspace{-1.7mm}
\section{Problem Characterization}
\vspace{-1.7mm}
\label{approach}
Sentiment classification in online social sites faces many challenges such as dealing with unstructured text and noisy user input, and mapping sentiment to objects or entities~\cite{liu2011opinion}. Beyond these, sentiment classification of news comments brings additional challenges, i.e., a variety of domains (e.g., politics, sports, and entertainment), lack of use of important sentiment clues (e.g., no use of emoticons), and the use of rhetorical questions. These additional, less studied challenges give rise to the unique design of our model.

The main tasks of sentiment classification of news comments are: (1) extracting entities from news comments, and (2) identifying users' sentiments about the extracted entities. Although both tasks have their own particular challenges, the second task is central to our study. To extract entities from news comments, we use the Stanford Named Entity Recognizer (SNER). SNER typically identifies three types of entities: person, place, and organization. More precisely, our problem can be formulated as follows.

{\bf Problem Formulation:} Given a comment and an entity, classify the sentiment expressed in the comment about that entity as: positive, negative or neutral/irrelevant. 

To address this problem, we decompose it into two parts. First, we link the target entity with its sentiment context. Specifically, when multiple entities are present in a comment, each entity must be linked to its own context, i.e., the words/phrases in the comment that are related to the entity. This is necessary since entities in a comment may have different sentiments or some entities may not have any sentiment at all associated with them (as illustrated below).

\noindent{\bf Example 5:} {\em In Ohio, voting for} Romney {\em who said he would let} GM {\em and} Chrysler {\em go bankrupt is like  paying a guy to rebuild your house that he burned down.}

Here, the sentiment is negative for Romney. However, GM and Chrysler do not have any sentiment.

Second, after entities are linked to their contexts, we identify the sentiment for an entity to be positive, negative or neutral, based on the sentiment of its context. 

\subsection{Extracting the Context of an Entity}
\vspace{-1.7mm}
The context of an entity contains the words, phrases or sentences that refer to the entity. We use several heuristics to extract the contexts. Following are the three main modules of our context extraction algorithm:

1. \textbf{Preprocessing}, where the number of entities in a comment is checked. For single entity comments, the entire comment is taken as the context for the entity. If a comment contains multiple entities, it is segmented into sentences and is given as input to the anaphora resolution module.

2. \textbf{Anaphora Resolution}: We use a rule based approach to anaphora resolution. We check the type of entity: PERSON (\textbf{P}) vs. NON-PERSON (\textbf{NP}) and assign sentences to the context of the entity if they have explicit mentions of that entity or {\em \textbf{compatible}} anaphoric references. For example, pronouns such as \emph{he, she, her, him} can only be used to refer to a \textbf{P} entity, whereas \emph{they, their, them} can be used to refer to both \textbf{P} and \textbf{NP} entities and {\em it} can only be used for \textbf{NP} entities. If a sentence does not have references to any entity, then it is added to the context of all the entities. Also, if a sentence has explicit mentions of multiple entities, then it is given as input to the local context extraction module.

3. \textbf{Local Context Extraction}: If entities occur in clauses that are connected with ``but'' (in the sentence), then the respective clauses are returned as local contexts for the entities. If the sentence contains a comparison between entities, then it is split at
the comparative term (adjective or adverb), with the comparative term added to the left part, and the two parts are returned as local contexts for the respective entities. If none of the two conditions is satisfied, then a window of $\pm 3$ tokens around entities is taken as their local context.

\vspace{-1.5mm}
\subsection{Identifying the Sentiment of Contexts}
\vspace{-2mm}
After obtaining the contexts of entities, we classify their sentiment into {\em positive}, {\em negative} or {\em neutral} sentiment classes. We model the task of identifying sentiment as two step classification. In the first step, we classify the context of an entity into {\em polar} versus {\em neutral} sentiment classes. Next, we classify the polar entities into {\em positive} or {\em negative} sentiment classes. Next, we describe the features used in our classification models and our reasoning behind using them.

\subsubsection{Neutral vs. Polar Classification}
\vspace{-1.8mm}
As already discussed, comments posted on news sites contain entities that are irrelevant with respect to sentiment analysis (see Example 1 in Section~\ref{intro}). These entities have no sentiment associated with them and are filtered out before conducting sentiment classification of comments. We address this problem by classifying entities as polar vs. neutral. Irrelevant entities are classified as neutral. Generally, content features and lexicon features form the basis of polar vs. neutral classification. However, in our data, we find some other interesting properties (specific to entities) that can be very helpful in identifying neutral and polar entities. For example, an entity that is a subject or direct object (of the subject) in a comment is more likely to be polar than an entity that is a prepositional object. Also, an entity of the type {\em person} is more likely to be polar than an entity that is of {\em non-person} type. Let us consider the following examples:

\noindent{\bf Example 9:} Bush {\em didn't blame anyone for trashing the} White House{\em , the 2001 recession, or for the 3 major attacks on} America.

\noindent{\bf Example 10:} Obama {\em stole 716 billion dollars we paid into} medicare.

In {\bf Example 9}, {Bush} is the subject, White House is the direct object and {America} is the prepositional object. In {\bf Example 10}, {Obama} is the subject, {Medicare} is the prepositional object. As we see, Obama and Bush are polar, whereas America, White House and Medicare are neutral.

Based on this reasoning, we extract the following features for all entities in a comment:

\textbf{IsPerson}: If the entity is of {\em person} type (1 if yes, 0 otherwise). To compute this feature, we look at the entity type output by SNER.

	\textbf{IsSubjObj}: If the entity is the subject, direct object, prepositional object or none of the three. (3 if subject, 2 if direct object, 1 if prepositional object, 0 otherwise). To compute this feature, we check if the entity has the following dependencies in the dependency tree: {\em nsubj } and {\em nsubjpass} (nominal subject and nominal subjective passive resp.), {\em dobj} (direct object) and {\em pobj} (prepositional object).
	
	 \textbf{HasClues}: If there are any polarity clues in the context of the entity, as detailed in Section \ref{sec:polar} (1 if yes, 0 otherwise). 
	
{\bf SentiPos:} This feature is calculated from the positive sentiment score given by the SentiStrength algorithm~\cite{thelwall2012sentiment} (0 if the score is 1, 1 otherwise) (we explain the scores output by SentiStrength in the following section).

{\bf SentiNeg:} This feature is calculated from the negative score given by the SentiStrength algorithm~\cite{thelwall2012sentiment} (0 if the score is -1, -1 otherwise).

\subsubsection{Positive vs. Negative Classification}
\label{sec:polar}
After obtaining the polar entities, we classify the sentiment about those entities into positive or negative sentiment classes. We use the following features for the positive-negative classification.

{\bf (a) Polarity Clues:}
Polarity clues are the words, phrases, or symbols used to express polarity of opinions/emotions. They have been used extensively in sentiment analysis~\cite{sentiWords,unsup1}. We use the subjectivity lexicon from MPQA corpus developed by Wiebe et al.~\cite{mpqa} to get the polarity clues. The lexicon contains $2006$ positive clues, $4713$ negative clues and $572$ neutral subjectivity clues. We extract three features {NumPos}, {NumNeg}, and {PosVsNeg} from the context of an entity. {NumPos} and {NumNeg} are the number of positive and negative polarity clues in the context, respectively. {PosVsNeg} is the number of positive divided by the number of negative polarity clues, i.e., (NumPos+1)/(NumNeg+1).

The following rules are used to count the polarity clues:

{\bf{\em Rule 1: Negation:}} If a polarity clue is connected to a negation word (i.e., they co-occur in a window of $3$ tokens), we reverse its polarity. If a neutral subjectivity clue is connected to a negation word, then its polarity is taken as negative. For example, {\em believe} is a subjectivity clue with prior polarity neutral, but if used with a negation (e.g. I do not believe or I cannot believe) expresses negative sentiment. We use a list of $50$ negation words.

{\bf{\em Rule 2: Quotes:}} Users often put polarity clues in quotes or in a quoted phrase to mean entirely opposite sentiment as compared to the sentiment expressed by the clue. If a polarity clue is in a quoted phrase then we reverse its polarity. Let us consider this example. 

\noindent{\bf Example 11:} The Republican party {\em also faces a steep climb with the} ``sane people'' {\em demographic.}

Here, the clue {\em sane} is in a quoted phrase {\em sane people}. The prior polarity of {\em sane} is positive. However, here, it is used to express negative sentiment about the Republican party.

{\bf{\em  Rule 3: ``but'' rule:}} Usually, sentiment expressed in clauses connected with ``but'' have opposite polarities. We take into account this property, while aggregating polarity clues for the entities. If clauses containing two entities are connected with ``but'' and there are explicit polarity clues in the context of only one of the entities, then we increase the count of the clue of opposite polarity for the other entity. 

\noindent{\bf Example 12:} {\em Read how} Bush {\em tried to control the financial situation with new regulations}, {\bf but} democrats  {\em blocked him.} Democrats {\em are pathetic, greedy liars.}

Here, Bush and Democrats occur in clauses connected with ``but'' and have opposite sentiment. For democrats, there are explicit negative clues (pathetic, greedy) but we do not have explicit polarity clues for Bush. In this case, we take the value of $NumPos$ feature for Bush as $1$.

{\bf {\em Rule 4: Comparatives:}} If two entities are present in a comparative clause and one of the entities does not have an explicit polarity clue (in its context), then for that entity we increment the number of the opposite polarity clue. We identify two most common types of comparatives: adjectival comparatives and adverbial comparatives. We look for {\em JJR} and {\em RBR} part-of-speech tags between entities to identify comparative adjectives and comparative adverbs, respectively. Let us consider this example:

\noindent{\bf Example 13:} {\em The} samsung {\em galaxys' are way better than all the mobile products} apple {\em puts out}.

Here, Apple has a negative sentiment but does not have any explicit polarity clue in its context. Using the rule, we take the value of $NumNeg$feature for Apple as $1$.

{\bf (b) Punctuation Marks:}
It is a common practice in online social media to use punctuation marks to express sentiments. We look for the presence of two punctuation marks: question and exclamation marks in the context of an entity. We calculate two punctuation features for a context: IsQuestion (presence or absence of a question mark), IsExclam (presence or absence of an exclamation mark).

{\bf (c) Sentiment Strength:}
These features capture the strength of the sentiments expressed in comments. We used the SentiStrength algorithm~\cite{thelwall2012sentiment} to compute these features. The algorithm is specifically designed to calculate sentiment strength of short informal texts in online social media. For a piece of text, the algorithm computes two integral scores, one in the range of +1 (neutral) to +5 (highly positive) that is expressive of the positive sentiment strength of the text and another in the range -1 (neutral) to -5 (highly negative) for negative sentiment strength. A score of +1 and -1 for a text means that the text is neutral or has no sentiment. Using SentiStrength, we compute three features: PosStrength (positive sentiment score), NegStrength (negative sentiment score) and PosVsNegStrength (PosStrength divided by NegStrength).

{\bf (d) Comment-specific features:}
These features capture clues that are specific to news comments. Users often use rhetoric in their comments to express a negative sentiment about an entity. They begin their comments by writing rhetorical questions and/or asking rhetorical questions about entities. Rhetorical questions are those that are not asked for the purpose of obtaining answers or information, but rather to make a point effectively\footnote{http://en.wikipedia.org/wiki/Rhetorical\_question}. Examples of rhetorical questions are: {\em Where is my vote?}, {\em Can't you do anything right?} Let us consider the  following examples:

\noindent{\bf Example 14:} {\em PLANS? What Plans?} Obama {\em has no plans for his second term.}

\noindent{\bf Example 15:} {\em So now the} Associated Press {\em has to correct their own corrections?}

These examples express implicit negative sentiment about Obama and Associated Press, without using explicit negative polarity clues. To capture these rhetorics, we design two binary features: IsFirstQues and IsEnQues. IsFirstQues checks whether the first sentence in the context of an entity is question or not. IsEnQues checks if an entity is present in a question sentence. To identify question sentences, we check for the presence of 5W1H words and question marks.

\section{Experiments}
\label{experiments}

\subsection{Data Description}
\begin{figure}[htb]
\centering
{\includegraphics[scale=0.245]{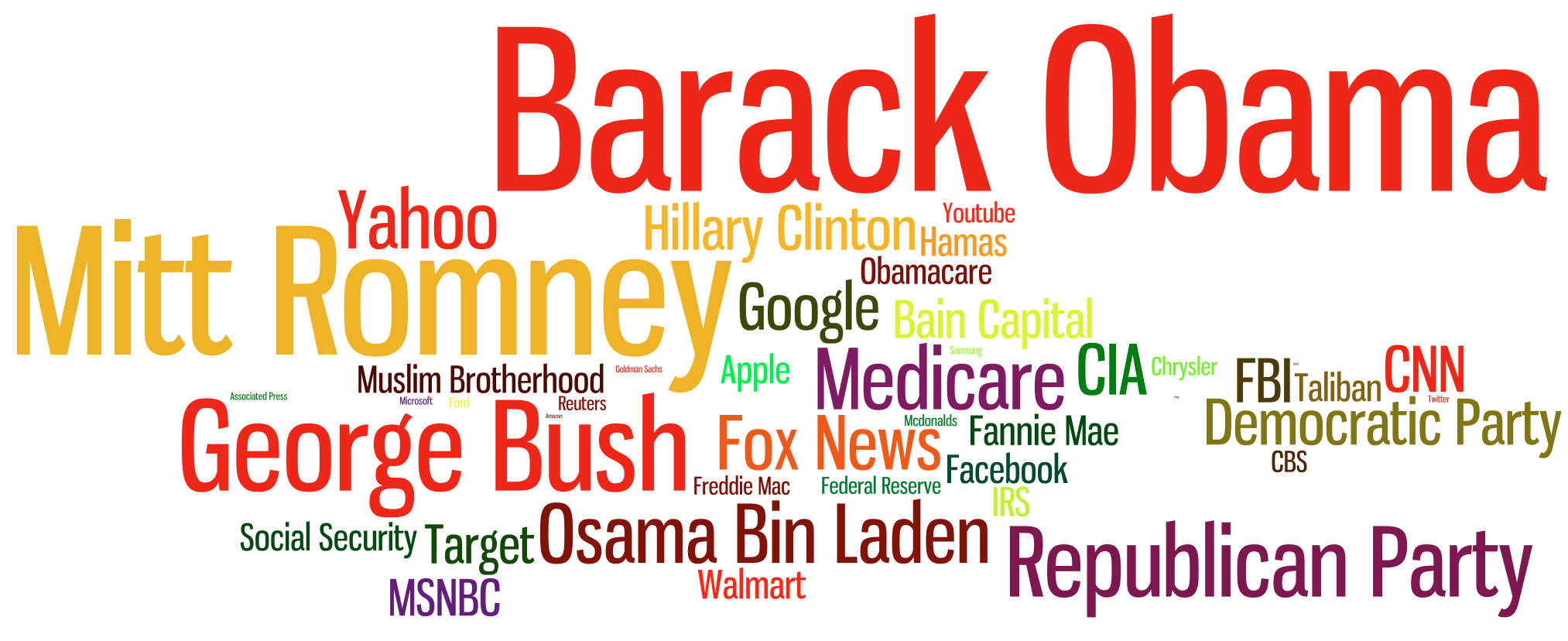}}
\caption{Word cloud of the $43$ entities used in sampling.} 
\label{fig-cc}
\end{figure}

Since there is no annotated dataset for sentiment classification of online news comments, we prepared our own dataset. We randomly sampled comments for annotation that satisfied certain constraints to ensure quality and diversity of the dataset. 
We, first, marked all the comments with the entities present in them and ranked the entities according to their comment frequencies. From the ranked list, we selected $43$ entities to consider. These entities covered areas such as politics (e.g., Obama, Romney), software (e.g., Google, Microsoft), online retail (e.g., Walmart, Ebay), hardware (e.g., Samsung, Apple), and insurance (e.g., Medicare, Obamacare), among others. The entities were selected based on their popularity as well as their relevance from the point of view of user targeting. Figure~\ref{fig-cc} shows a ``word cloud'' of the $43$ entities. The larger the entity, the more frequent it is in the news comments. As we see, {\em Walmart} has a much smaller comment frequency compared with other entities such as {\em Barack Obama}, however, it is important due to its commercial nature and relevance to ad targeting. 


We sampled $526$ comments such that all the comments have at least one of the $43$ entities and all the entities have approximately equal number of sampled comments. We then marked the three most important entities in each comment, obtaining $941$ instances. For a comment, the number of instances is equal to the number of entities marked for that comment. 
Each instance was annotated by two annotators. For each instance, the annotators were asked to identify sentiments expressed in that instance (as negative, neutral or positive). The agreement between the annotators was $90\%$. For the remaining $10\%$, a third annotator was asked to select between the two annotations of the two original annotators. Given this annotation scheme, we obtained $632$ negative instances, $151$ positive instances and $158$ neutral instances. Also, $41$ comments are neutral, i.e., all the entities present in them have neutral sentiment, $184$ comments contain polar as well as neutral entities and $301$ comments have only polar entities. We call the comments that have both polar and neutral entities as pseudo-polar comments.  

\subsection{Experimental Setting}
\label{sec:setting}
We conducted sentiment classification experiments using various supervised machine learning algorithms implemented in the Weka data mining toolkit \cite{hall2009weka}. For neutral-polar classification, Logistic Regression gave the best performance, whereas for the positive-negative classification, Naive Bayes outperformed other supervised methods. To evaluate the performance of our classifiers, we report precision, recall and F-1 score, all macro averaged across $10$ folds in a cross validation setting. 

For neutral-polar classification, we use neutral and pseudo-polar comments. After segmenting comments into contexts of entities, we obtain a total of $345$ instances ($158$ neutral and $187$ polar) from $225$ comments. As explained in Section~\ref{approach}, an instance for classification is a context of an entity present in a comment. Since a comment may have multiple entities and, hence, multiple contexts, we can obtain more instances than the total number of comments. For positive-negative classification, we use polar and pseudo-polar comments (a total of $485$ comments). Neutral entities from the pseudo-polar comments are not considered in positive-negative classification.
 
{\bf Baselines:} We compare our sentiment classifiers with the following three baselines:

{\bf 1. Bag-of-words and POS tags} \cite{Jiang2011,supervised1,McDonald2007}: We use the words in the context of an entity and the part-of-speech tags of those words as features for classification and experiment with two settings: 1) BoW, in which only word frequencies are used as features, 2) BoW+POS, in which both word frequencies and their POS tags are used as features. We use Multinomial Na\"ive Bayes for these models.

{\bf 2. SentiStrength}: SentiStrength is a state-of-the-art tool for sentiment analysis of short informal texts posted on online social media. 
 We use the following two settings for turning SentiStrength into a sentiment classifier:

	\begin{enumerate}[(a)]
		\item {\bf SentiStrength scores as features}: We use the two scores (positive and negative) output by SentiStrength as features for sentiment classification. 
		\item {\bf SentiStrength scores as rules}: We use the two scores directly as rules for making an inference about the sentiment of a context. For neutral-polar classification, a score of +1 and -1 implies that the text is neutral, and polar otherwise. For the positive-negative classification, a context is positive if its positive sentiment score is greater than its negative sentiment score and similarly for inferring a negative sentiment. For example, a score of +3 and -2 implies positive polarity and a score of +2 and -3 implies negative polarity. If both scores are equal for a context, we randomly assign the context to one class or the other.
		\end{enumerate}
		
{\bf 3. LexiconRuleBased} \cite{ding2008holistic,meng2012entity}: We compute the sentiment for an entity $e$ in a comment $C$ by calculating the following score:
		\begin{equation}
			score(e,C)= \sum_{s\in C}{\left (  \sum_{w_i:w_i\in s \cap w_i\in L}{\frac{w_i.SO}{d(e,w_i)}}\right )}
		\label{eq:base}
		\end{equation}
	where  $s$ is a sentence in $C$, $w_i$ is a polarity word in $s$, $L$ is the polarity lexicon, $w_i.SO$ is the sentiment orientation of $W_i$ (1 if positive, -1 if negative) and $d(e,w_i)$ is the distance between the polarity word $w_i$ and $e$ in $s$. The denominator down-weights the sentiment orientation of polarity words that are far from the entity. The sentiment is positive if the score is greater than zero, negative if the score is less than zero and neutral otherwise. 
For positive-negative classification, if we obtain a zero score, we assign the entity randomly to the positive or the negative class.
	
{\bf 4. Naive context extraction:} To evaluate our context extraction algorithm, we compare it against a simple method for extracting entity contexts. For this method, we extract entity contexts using a simple scheme. We add the entire sentence to the contexts of all the entities present in it. If a sentence does not contain any entity, then we add it to the context of all the entities in the comment. All other classification settings (features and classifiers) remain same.

\begin{table}[ht]
\begin{small}
\begin{centering}
	\begin{tabular}{@{} p{1.6 in} c c c @{}}
		\toprule
		{\bf Model}&{\bf Pr.}&{\bf Re.}&{\bf F-1} \\
		\midrule
		lexiconRuleBased &$0.515$ & $0.420$& $0.463$\\
		SentiStrength (rule) & $0.527$ & $0.502$ & $0.466$\\	
		SentiStrength (features) & $0.506$ & $0.510$ & $0.507$ \\	
		BoW& $0.553$ & $0.557$ & $0.553$  \\	
		BoW+POS & $0.565$ & $0.565$ & $0.565$ \\	
		\midrule
		NaiveContextExtraction & $0.646$ & $0.645$ & $0.645$ \\	
		\midrule
		Proposed model$-$IsPerson & $0.575$ &$0.574$ & $0.574$ \\
		Proposed model$-$IsSubjObj  & $0.680$ &$0.643$ & $0.636$\\
		Proposed model $-$HasClues  & $0.667$ & $0.664$ & $ 0.664$ \\
		 Proposed model$-$SentiStrnth & $0.667$&$0.664$&$0.664$ \\
		 \midrule
		{\bf Proposed model} & ${\bf 0.671}$ &  ${\bf 0.670}$&  ${\bf 0.670}$ \\	
		\bottomrule
	\end{tabular}
	\caption{Neutral-polar Classification results.}
	\label{neutralClassResults}
\end{centering}
\end{small}
\end{table}
\subsection{Classification Results}
\subsubsection{Neutral-polar Classification}
Table~\ref{neutralClassResults} shows the results of neutral-polar classification. The first five rows show the results of the baseline models, whereas the subsequent four rows show the results of models built by removing only one feature at a time from the proposed model. The last row shows the result of the proposed model. As can be seen from the table, the proposed model outperforms all the baselines and using all the features gives the best performance with F-1 score of $0.67$. We see that the LexiconRuleBased method and SentiStrength (rule) are the worst performing models with F-1 scores of $0.463$ and $0.466$, respectively, followed by SentiStrength (features) with an F-1 score of $507$. This can be attributed to the fact that SentiStrength is trained on {\em online social media data}, which is significantly different from comments data. For example, one of the features used by SentiStrength for detecting sentiment is the presence of emoticons, which are generally not present in news comments.  
Similarly, we see that the BoW model performs the third worst with an F-1 score of $0.553$. Adding part-of-speech tags to BoW improves the performance to an F-1 score of $0.565$. Note that BoWs generally perform better in other sentiment classification tasks in domains such as Twitter~\cite{Jiang2011} and product reviews~\cite{supervised1,McDonald2007} compared with BoWs in our domain. A possible reason could be the presence of implicit sentiment in the form of rhetorical questions, sarcasm, etc., where users do not use explicit sentiment words, and hence, there are less patterns of words and common POS tags that are generally used to express sentiment and subjectivity (e.g., adjectives, adverbs and common nouns) to learn the models.  Also, we see that our method outperforms NaiveContextExtraction. 

Next, we discuss the impact of removing different features from our model. We see that removing the IsPerson feature decreases the F-1 score to $0.574$ and when IsSubjObj is removed the performance drops to $0.636$. The removal of HasClues and SentiStrength features (sentiPos and sentiNeg) has a similar impact on the performance (however, not as big as the removal of IsPerson or IsSubjObj), resulting in an F-1 score equal to $0.664$, in both cases. We see that removing IsPerson feature has the highest negative impact on the performance, followed by IsSubjObj feature and HasClues and SentiStrength features. This observation is consistent with the feature ranking using Information Gain (IG) \cite{yang1996feature} as output by Weka. The following is the IG-based feature ranking: IsPerson $>$ IsSubjObj $>$ SentiNeg $>$ HasClues $>$ SentiPos. The features on the right side of $>$ have higher rank than those on the left side of $>$. We see that the two proposed non-lexical features, IsPerson and IsSubjObj, are more informative than HasClues and SentiStrength features that are based on lexical properties of comments. This suggests that in comments, entity type ({\em person} or {\em non-person}) and its grammatical role in the comment (subject, direct object or prepositional object) are highly informative clues/features for polarity.

\begin{table}[t]
\begin{small}
\begin{centering}
	\begin{tabular}{@{} p{1.6 in} c c c @{}}
		\toprule
		{\bf Model}&{\bf Pr.}&{\bf Re.}&{\bf F-1} \\
		\midrule
		lexiconRuleBased & 0.227&0.432&0.298\\
		SentiStrength(rule) & $0.44$&$0.462$ &$0.452$\\
		SentiStrength(feature) & $0.599$&$0.77$ &$0.674$\\
		BoW & $0.637$&$0.687$ &$0.659$\\
		BoW+POS & $0.653$&$0.693$ &$0.665$\\
		\midrule	
		NaiveContextExtraction &$0.472$ & $0.492$& $0.491$\\
		\midrule
		Propsed model$-$CSF  & $0.673$&$0.715$&$0.687$\\
		{\bf Proposed model}  & ${\bf 0.678}$&${\bf 0.717}$&${\bf 0.700}$ \\		 
		
		\bottomrule
	\end{tabular}
	\caption{\small{Positive-negative classification results.}}
	\label{polar}
\end{centering}
\end{small}
\end{table}

\subsubsection{Positive-negative Classification}
\label{polarClass}
Table~\ref{polar} shows the results of positive-negative classification experiments. The first five rows present the results of the four baseline classification models, whereas the next two rows show the results of the proposed model without and with comment-specific features, denoted by Proposed model$-$CSF and Proposed model, respectively. 
As can be seen, the LexicalRuleBased method is the worst performing model in this setting  with an F-1 score of $0.298$, followed by SentiStrength (rule), BoW and SentiStrength (features) with F-1 scores of $0.452$, $0.659$ and $0.674$, respectively. POS tags improve the F-1 score of BoW model from $0.659$ to $0.665$. The proposed model outperforms all the baselines, having an F-1 score of $0.7$. To see the effect of comment-specific features on the positive-negative classification, we experimented with the proposed model without the comment-specific features. We see that adding comment-specific features improves the F-1 score of the model from $0.687$ to $0.7$. 

To analyze the importance of different features, we ranked them using Information Gain \cite{yang1996feature} and obtained the following feature ranking: NumNeg $>$  PosVsNeg $>$ NegStrnth $>$ IsQuesMark  $>$  IsEnQues $>$ PosStrnth  $>$ IsQuesFirst $>$ IsExclaim $>$ NumPos $>$ PosVsNegStrnth. We see that features related to positive sentiment (PosStrnth and NumPos) are ranked lower than NumNeg and NegStrnth features. One potential reason for this is that users generally express negative sentiments {\em more explicitly} than positive sentiments, and hence, the presence of significantly more negative patterns to learn as compared to the positive ones.

\vspace{-1.8mm}
\section{Conclusion and Future Work}
\label{sec:Conclusion}
\vspace{-1.8mm}
In this paper, we studied the problem of identifying users' sentiments towards individual entities referenced in comments on news articles. We identified several challenges to this problem and proposed solutions to address them. In particular, we designed an algorithm to extract the context of entities in comments, proposed {\em novel} non-lexical features for neutral-polar classification, and comment specific features for polarity classification. Our methods outperformed strong baselines for sentiment classification. Interesting directions for future work include: (1) using priors on users based on their comments on (particular or all) entities, e.g., a user could be pessimistic or cynical towards all entities; (2) training mixture of specialized classifiers for the domains covered by a news site, e.g., political, sports, technology, and entertainment. We believe generally people become more sarcastic when they discuss politics.

\begin{footnotesize}
\bibliographystyle{aaai}
\bibliography{References}
\end{footnotesize}

\end{document}